\documentclass[review]{fcs}
\usepackage{layout}
\title{Position: How can Graphs Help Large Language Models?}

\author[1,*]{Xiyuan Wang}
\author[1,*]{Yi Hu}
\author[1,*]{Yanbo Wang}
\author[2]{Chuan Shi}
\author[1,+]{Muhan Zhang}
\address[1]{Institute for Artificial Intelligence, Peking University, Beijing 100871, China}
\address[2]{School of Computer Sciences, Beijing University of Posts and Telecommunications, Beijing 100876, China}
\corremail{muhan@pku.edu.cn}
\fcssetup{
  received = {month dd, yyyy},
  accepted = {month dd, yyyy},
}

\begin{abstract}
With the rapid advancement of large language models (LLMs), classic graph learning tasks have greatly benefited from LLMs, including improved encoding of textual features, more efficient construction of graphs from text, and enhanced reasoning over knowledge graphs. In this paper, we ask a complementary question: “How can graphs help LLMs?” We address this question from three perspectives: 1) graphs provide an up-to-date knowledge source that helps reduce LLM hallucinations, 2) graph-based prompting techniques—such as Chain-of-Thought (CoT), Tree-of-Thought (ToT), and Graph-of-Thought (GoT)—enhance LLM reasoning capabilities, and 3) integrating graphs into LLMs improves their understanding of structured data, expanding their applicability to domains such as e-commerce, code, and relational databases (RDBs). 
We further outlook some future directions including designing sparse LLM architectures based on graphs and brain-inspired memory systems.
\end{abstract}

\begin{document}
\section{Introduction}
While large language models (LLMs) have significantly advanced graph learning—enriching text representations, streamlining the construction of graphs from text, and enhancing reasoning over knowledge graphs~\cite{li2023survey, chen2024exploring, wang2025graph, jin2024large, ren2024survey,zhu2024llms}—a compelling and complementary question arises: How can graphs, in turn, empower LLMs?

{Recent surveys have explored the general landscape of this intersection. Pan et al.~\cite{pan2024unifying} and Kau et al.~\cite{kau2024combining} systematically categorize the architectural symbiosis between LLMs and Knowledge Graphs. While these works primarily dissect the \textbf{architectural modes} of integration for Knowledge Graphs, our work specifically isolates the ``Graph $\to$ LLM'' perspective. We distinguish our survey by analyzing the \textbf{functional utility} of graph structures rather than viewing integration solely through an architectural lens.}

\begin{figure*}[h!tb]
    \centering
    \includegraphics[width=0.8\textwidth]{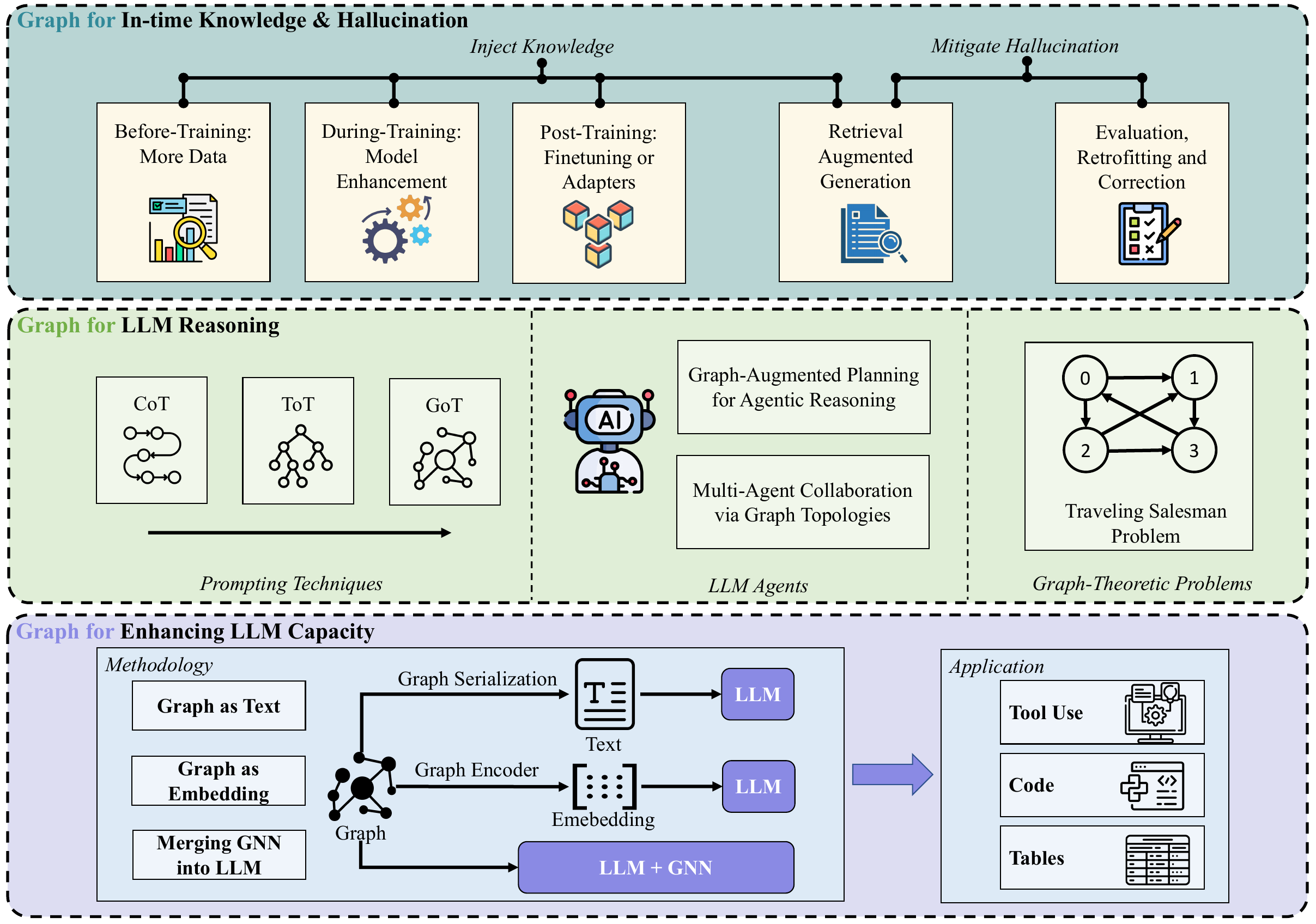}
    \caption{A Taxonomy of How Graphs Empower Large Language Models}
    \label{fig:taxonomy}
\end{figure*}

\begin{figure*}[ht]
    \centering
    \includegraphics[width=0.8\textwidth]{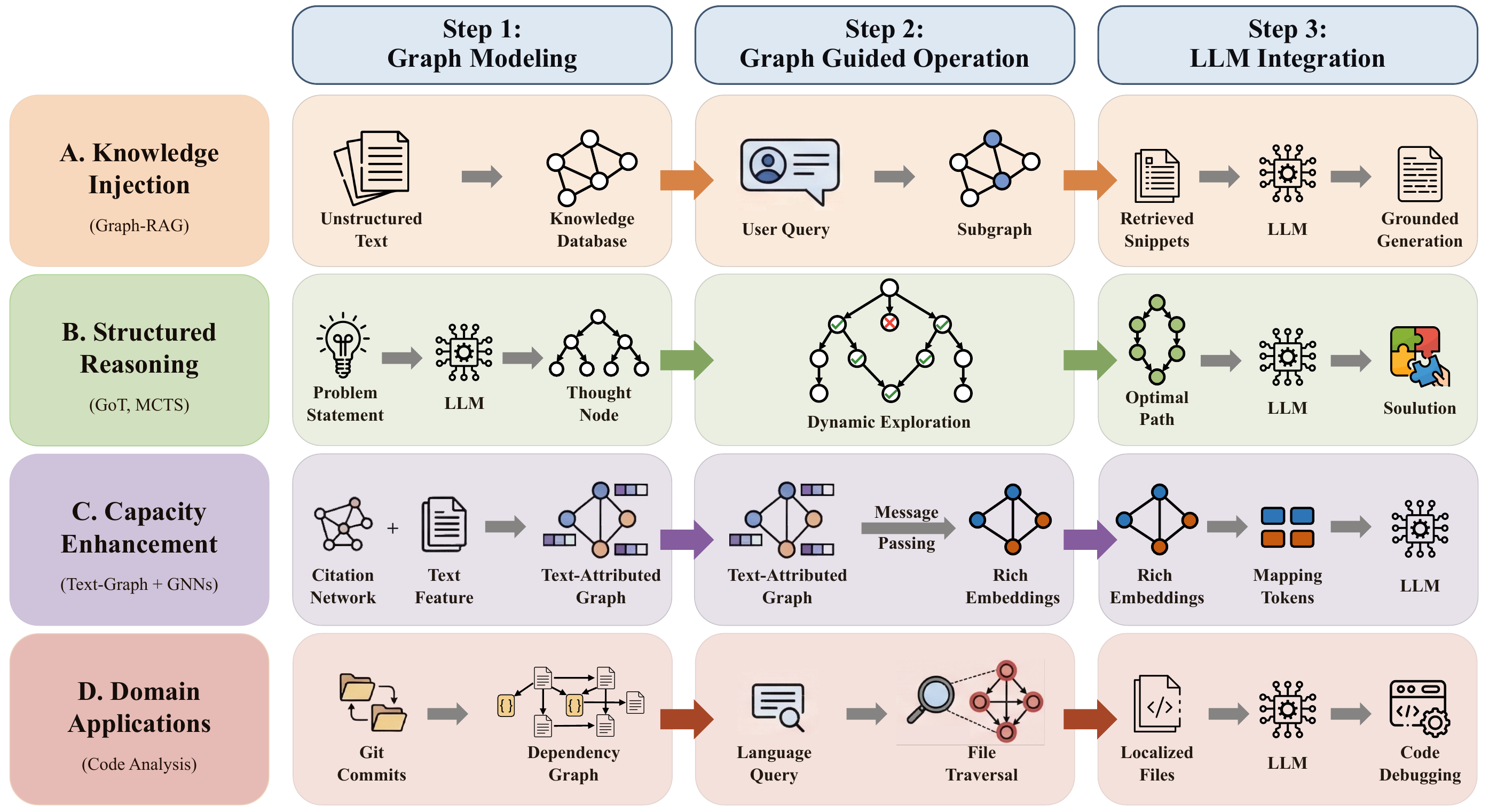}
    \caption{An overview of the operational steps across various Graph-for-LLM paradigms. Using representative scenarios for each category (e.g., GraphRAG for knowledge injection, or code dependency parsing for domain applications), this figure provides a simplified illustration of how raw data or reasoning states are translated into graph structures, manipulated via graph-specific operations, and finally fed into the LLM to execute or enhance a specific task.}
    \label{fig:illustration}
\end{figure*}

We systematically investigate how graphs serve as a powerful infrastructure to mitigate these limitations and broaden LLM applicability. {Specifically, as summarized by the taxonomy in Figure~\ref{fig:taxonomy},} this survey is structured around three key perspectives: 1) \textbf{Graphs for Real-Time Knowledge \& Hallucination Reduction:} Leveraging the structured and updatable nature of knowledge graphs to provide factual grounding. 2) \textbf{Graphs for LLM Reasoning:} Employing graph-based prompting paradigms—such as Chain-of-Thought (CoT), Tree-of-Thought (ToT), and Graph-of-Thought (GoT)—to organize and enhance complex reasoning processes. 3) \textbf{Graphs for Enhancing LLM Capacity:} Integrating graph structural information into LLMs to improve their comprehension and performance in structured-data domains like web navigation, code analysis, and relational databases. {To aid in navigating these diverse methodologies, Figure~\ref{fig:illustration} provides a conceptual overview of the typical operational steps across various Graph-for-LLM paradigms. By highlighting specific representative workflows, such as GraphRAG for knowledge injection or code analysis for domain applications, it illustrates how raw data or reasoning states are generally translated into graph structures, manipulated via graph-specific operations, and finally fed into the LLM.} Furthermore, we outline \textbf{promising future research directions}, such as designing sparse LLM architectures that leverage graph-based computation and systems inspired by human memory mechanisms.

\section{Graph for In-time Knowledge \& Hallucination Reduction} 




The integration of structured knowledge from knowledge graphs (KGs) into pre-trained language models (PLMs) represents a significant effort to overcome their inherent limitations. While proficient at processing text, LLMs often lack deep, grounded factual understanding, leading to issues like ``hallucination''. KGs, with their explicit, structured, and easily updatable nature, offer a powerful solution.

These works can be broadly categorized into five strategies based on when and how knowledge is integrated. Before-Training Enhancement is a data-centric approach that modifies the training data itself, weaving facts directly into the text so a standard model can learn them. During-Training Enhancement is an architecture-centric approach that modifies the model's internal structure or core training objectives to build a new, fundamentally more knowledgeable foundation model. Post-Training Enhancement is an adaptation-centric approach, taking an already pre-trained model and specializing it for knowledge-intensive tasks using efficient methods like fine-tuning or lightweight adapters. Retrieval-Augmented Generation (RAG) is an inference-time approach that treats the KG as an external, ``open-book'' resource, retrieving relevant facts on-the-fly to provide context for the LLM's response. Finally, Post-Generation Hallucination Mitigation is a specialized category of methods that operate on the LLM's output to evaluate, correct, and retrofit its generated text for factual accuracy.

\subsection{Before-Training Enhancement: Improving the Learning Data}

The most direct way to make language models more knowledgeable is to enrich the pre-train data with required facts. One pioneering method, introduced in K-Bert~\cite{liu2020k}, injects knowledge graph facts directly into sentences to create a richer, tree-like structure for the model to process. A different strategy, proposed by Xiong et al.~\cite{xiong2019pretrained}, trains the model to develop a better sense of factual consistency by swapping correct entities in a sentence with incorrect ones. Some works focus on creating unified representations. Sun et al.~\cite{sun2020colake} builds a ``word-knowledge graph'' to combine both language and knowledge contexts into a single structure. To help models learn about more important concepts, Shen et al.~\cite{shen2020exploiting} propose to use a KG to guide the model's training by intelligently masking important entities. Yet Zhang et al.~\cite{zhang2022dkplm} focuses on enriching the training data for ``long-tail'' entities that are important but appear infrequently.

\subsection{During-Training Enhancement: Building a More Knowledgeable Model}

A more fundamental approach is to modify the model's internal architecture or its core training process from the ground up, aiming to build a new foundation model that is inherently better at understanding and integrating knowledge.

This can be done by adding a dedicated module for processing graph information. For instance, ERNIE~\cite{zhang2019ernie} uses a separate encoder for entities, while Bert-MK~\cite{he2019integrating} learns from entire subgraphs of related facts, making it highly effective for specialized domains like medicine. A more advanced technique used by Yasunaga et al.~\cite{yasunaga2022deep} train the model on understanding text and graph structures at once.

Other works introduce components to fuse information. KLMo~\cite{he2021klmo} uses a ``knowledge aggregator'' to connect text to the KG, while KnowBert~\cite{peters2019knowledge} inserts a special layer into a BERT model for real-time KG lookups. A more complex interaction proposed by Yu et al.~\cite{yu2022jaket} creates a partnership between the language and knowledge modules so they can improve each other. To make this intensive process more efficient, Trelm~\cite{yan2024trelm} identifies the specific parts of the model that store factual knowledge and focuses on updating only those, saving significant time and resources.

\subsection{Post-Training Enhancement: Specializing an Existing Model}
Given the high cost of pre-training, a practical alternative is to adapt an existing LLM for knowledge-intensive tasks. One way is to fine-tune the model on a specific task, augmenting it with a graph-reasoning module. Models such as ~GreaseLM~\cite{zhang2022greaselm} and JoinkLK~\cite{sun2021jointlk} use this approach to help an LLM reason over a KG for tasks like question answering. For generative tasks, GAP~\cite{colas2022gap} fuses graph-aware elements into an existing LM encoder to better capture the KG's topology without requiring additional pre-training tasks. An even more efficient method is to use small modules like adapters to plug into a frozen LLM with new knowledge~\cite{wang2020k, tian2024kg}.

\subsection{Retrieval-Augmented Generation (RAG): Providing an Open Book at Test Time}
A more flexible approach to knowledge utilization is RAG. Edge et al.~\cite{edge2024local} introduced the GraphRAG system, which treats the knowledge graph (KG) as an external, ``open-book'' resource. The method employs an LLM to construct a KG from a document corpus, then identifies and summarizes communities of entities to address broad, corpus-level questions. Subsequent works have extended this idea: LEGO~\cite{cao2024lego} performs fine-grained, modular analysis; LightRAG~\cite{guo2024lightrag} incorporates dual-level information with efficient operations; and hybrid systems such as ToG2~\cite{ma2024think} and HybridRAG~\cite{sarmah2024hybridrag} leverage retrieval from both KGs and plain-text documents. Other research, including G-Retriever~\cite{he2024g} and SubgraphRAG~\cite{li2024simple}, focuses on developing highly efficient methods for retrieving the most relevant and compact subgraph from a KG to answer a given question.

{\textbf{Scalability and Computational Challenges.} While integrating graphs enhances reasoning depth, it shifts significant computational burdens to both the pre-processing and test-time phases. First, construction overhead remains a primary bottleneck; generating graph indices with LLMs is resource-intensive, with Jain et al.~\cite{jain2025aligning} reporting a 10-20x increase in resource consumption compared to vector baselines. For instance, Edge et al.~\cite{edge2024local} found that indexing just 1 million tokens required 281 minutes with GPT-4. To mitigate this, recent works have shifted toward LLM-free construction; LinearRAG~\cite{zhuang2025linearrag}, for example, reduces indexing time by over 77\% by bypassing complex relation extraction. Second, dynamic maintenance presents a critical barrier to "real-time" RAG. Due to the high construction costs noted above (e.g., $\approx$4.7 hours for 1M tokens), most systems rely on static snapshots, necessitating full graph reconstruction for minor corpus updates. Emerging research addresses this rigidity through incremental update frameworks; notably, EraRAG~\cite{zhang2025erarag} employs locality-sensitive hashing to map new data to existing graph clusters, reducing update time by 95\% and making dynamic integration viable. Finally, inference latency is critical; complex subgraph identification strategies can introduce latencies exceeding 600 seconds on dense graphs~\cite{li2024simple}. Recent optimizations, such as SubgraphRAG's lightweight scoring ($\approx$12s)~\cite{li2024simple} and SubGCache's subgraph-level caching (e.g., 6.68x speedup)~\cite{zhu2025subgcache}, are essential for aligning graph reasoning with interactive requirements.}

\subsection{Mitigating Hallucinations: KG-based Evaluation, Retrofitting, and Correction}
A key application of KG integration methods is the mitigation of LLM hallucinations. As KGs provide verifiable, structured facts, they serve as effective tools for grounding outputs and correcting errors. Existing methods can be broadly grouped into two categories: (1) \emph{post-generation evaluation} for detecting errors, and (2) \emph{post-generation retrofitting or correction} for actively fixing them.

The first category, \textbf{Post-Generation Evaluation}, leverages KGs as ground truth to assess hallucinations after an LLM produces a response. This often requires converting the unstructured text output into a structured representation that can be systematically verified. For example, GraphEval~\cite{sansford2024grapheval} and GCA~\cite{fang2025zero} translate the LLM's generated text into a knowledge graph, then analyze it for internal inconsistencies or logical contradictions—strong indicators of hallucination. Liu et~al.~\cite{liu2024evaluating} take a different approach, using a large-scale KG to automatically generate millions of true/false questions. Instead of directly querying the target LLM for each question, which would be computationally expensive, they train a smaller, efficient ``judge model'' to estimate the LLM’s performance, enabling large-scale factuality assessment at a fraction of the cost.

The second category, \textbf{Post-Generation Retrofitting and Correction}, actively revises hallucinated content through multi-step, closed-loop processes. Frameworks such as KGR~\cite{guan2024mitigating}, Re-KGR~\cite{niu2024mitigating}, and TrustfulLLM~\cite{zhu2024trustful} implement a ``retrofitting'' or ``refinement'' loop: they extract key factual claims from the LLM’s draft, verify them against a trusted KG, and prompt the LLM to revise its output based on factual feedback. This iterative process aligns the final response more closely with the ground truth. Other methods address specific tasks; for instance, PGMR~\cite{sharma2025reducing} focuses on generating SPARQL queries, using a KG to retrieve correct database identifiers (URIs) and insert them into the LLM-generated query structure, thereby avoiding incorrect links. Finally, empirical work by Barkley~\cite{barkley2024investigating} shows that different prompting strategies and the integration of external tools, including KGs, can significantly reduce hallucination rates, reinforcing the value of this research direction.

\section{Graph for LLM Reasoning}
Graph-based prompting techniques have increasingly gained prominence in enhancing LLMs by providing more sophisticated structures for reasoning. These techniques employ graph topologies—such as chains, trees, and dynamic graphs—to support more flexible and complex thought processes. The evolution from basic Chain-of-Thought (CoT)~\cite{few-shot-cot, zero-shot-cot} to more intricate models like Tree-of-Thought (ToT)~\cite{tot} and Graph-of-Thought (GoT)~\cite{got} reflects this shift towards incorporating multiple reasoning paths and recursive thought structures. Additionally, research combining LLMs with Monte Carlo Tree Search (MCTS)~\cite{xot,wan2024alphazero,qi2024mutual,guan2025rstar,zhang2024rest,zhang2024accessing,tian2024toward} introduces dynamic exploration within a computational graph, further enriching the reasoning capabilities of LLMs. These approaches not only offer a broader scope for processing diverse types of reasoning but also enable LLMs to adaptively traverse and synthesize multiple possible solutions.

Graph-based structures are also proving valuable in LLM agent applications, particularly in enhancing agentic reasoning and multi-agent collaboration. Graph-augmented planning enables more efficient handling of complex tasks by structuring dependencies and subtasks, while various communication topologies in multi-agent systems enhance coordination, scalability, and efficiency. Through task-adaptive graphs and optimization techniques, these systems can refine how agents interact, manage workflows, and perform collective reasoning, addressing key limitations in sequential and parallel task management. These innovations underscore the power of graph-based structures in both individual and multi-agent LLM applications, facilitating more nuanced and scalable decision-making processes.

\subsection{Graph-based Prompting Techniques}
Prompting techniques for LLM reasoning have evolved from simple linear chains to complex graph-based structures. In Chain-of-Thought (CoT)~\cite{few-shot-cot,zero-shot-cot}, reasoning is modeled as a single chain of intermediate steps between input and output. Variants such as CoT-SC~\cite{cot-sc} introduce multiple independent chains to explore the space of possible solutions. Tree-of-Thought (ToT)~\cite{tot} generalizes this by allowing dynamic branching during reasoning, where each partial solution can expand into several new candidates based on heuristic or learned scoring. Both chain and tree structures are, in fact, specific types of graphs: chains are linear graphs, while trees are directed acyclic graphs with a single parent per node.

Graph-of-Thought (GoT)~\cite{got} further generalizes reasoning topologies by allowing arbitrary connectivity---nodes (thoughts) can have multiple parents and multiple children, supporting both branching and aggregation. This enables expressive reasoning patterns such as dynamic programming or analogical propagation, where intermediate sub-solutions can be composed and reused across multiple inference paths.

{Chain-based methods~\cite{few-shot-cot,zero-shot-cot} handle sequential reasoning well but struggle with complex, multi-step problems. Tree-based approaches~\cite{tot} address this by supporting structured problem decomposition and guided exploration through user-defined constraints. While computationally intensive, tree-based methods outperform chain-based ones in tasks with clear hierarchical structure, such as arithmetic and symbolic reasoning.
Graph-based approaches~\cite{got} further generalize tree structures by allowing non-hierarchical, interconnected reasoning, where nodes can influence each other dynamically. This boosts performance in multi-hop and arithmetic reasoning. 
However, these advanced architectural gains come at the cost of significant computational overhead.
\textbf{Empirical evidence from complex document merging tasks~\cite{got, xot} clearly illustrates this trade-off. While CoT achieves a score of 6.524 with a modest consumption of 3,153 tokens, moving to advanced structures leads to a significant increase in resource usage. Specifically, ToT achieves the highest performance (7.715) but requires 51,486 tokens—over 16 times the cost of CoT. GoT offers a more favorable trade-off, yielding a score of 7.559 with 27,685 tokens; yet, even GoT remains nearly 9 times more expensive than simple chain-based reasoning~\cite{xot}. }
Notably, GoT's performance gains are more pronounced on harder problems, whereas achieving similar results with ToT tends to be more costly~\cite{besta2025demystifying}. However, GoT's effectiveness relies on careful manual design, including problem structuring and prompt design, to fully realize its systematic problem-solving potential~\cite{chen2025unleashing}.}

Complementary to explicit graph prompting, Monte Carlo Tree Search (MCTS) offers a dynamic framework for LLM reasoning by treating the reasoning process as a computational graph where nodes represent intermediate states and edges capture transitions between them. In this paradigm, methods like~\cite{xot,wan2024alphazero,qi2024mutual,guan2025rstar,zhang2024rest,zhang2024accessing,tian2024toward} leverage the graph’s flexibility to explore, backtrack, and synthesize reasoning paths, enabling LLMs to decompose complex problems into manageable steps. Unlike chain or tree-based approaches, MCTS allows for adaptive exploration—sampling diverse trajectories, evaluating partial solutions, and iteratively refining the search space. This mirrors the expressive power of general graph-based reasoning, where thoughts can be dynamically connected, reused, or discarded based on their utility. By framing MCTS as a graph-guided search process, these methods enhance LLMs’ ability to navigate large, structured reasoning spaces efficiently.

Parallel to graph-structured prompting, another line of work models the reasoning process as a computation graph through program representations. In these approaches, the reasoning steps are expressed as structured programs. Prior efforts~\cite{gao2023pal,chen2022program,hu2023code,hu2024case,hu2025beyond} aim to model LLM reasoning as interpretable and executable computation graphs. Approaches such as~\cite{gao2023pal,chen2022program} enable LLMs to offload complex math and symbolic operations to external programmatic engines, thereby enhancing overall reasoning performance. In contrast, methods like~\cite{hu2023code,hu2024case,hu2025beyond} explore the model's intrinsic ability to reason with the aid of code, teaching LLMs to imitate a code interpreter and ``execute'' programs internally, without relying on external tools.

\subsection{Graph-based Applications in LLM Agents}

\paragraph{Graph-Augmented Planning for Agentic Reasoning} Recent work has explored the integration of graph structures into LLM-based planning to address the limitations of LLMs through sequential reasoning over structured task spaces. In complex scenarios involving both sequential and parallel steps, subtasks and their dependencies can be naturally modeled as task graphs. Graph-based prompting techniques~\cite{lin2024graph}  and hybrid architectures combining LLMs with GNNs~\cite{wu2024can} have been shown to enhance planning capabilities. Benchmarks such as WORFBENCH~\cite{qiao2025benchmarking} further reveal significant performance gaps between sequence-based and graph-based planning, highlighting the importance of explicit graph reasoning in agentic workflow generation.

\paragraph{Multi-Agent Collaboration via Graph Topologies} In multi-agent settings, graphs provide a natural representation for organizing communication, coordination, and role assignment among LLM-based agents. Prior research has adopted diverse communication topologies—--such as chains~\cite{few-shot-cot,zero-shot-cot}, trees~\cite{tot}, complete graphs~\cite{qian2025scaling}, random graphs~\cite{qian2025scaling}, and optimizable graphs~\cite{zhuge2024gptswarm,zhang2024cut,zhang2024g}---to structure inter-agent interactions. These topologies not only shape the flow of information but also impact scalability. Recent approaches~\cite{zhang2024g} model multi-agent systems as task-adaptive graph and optimize them based on task-specific context. Regarding efficiency, \cite{zhang2024cut} introduces a spatial-temporal graph pruning framework that identifies and removes redundant or uninformative communication, enhancing reasoning efficiency.

\subsection{Graph-Theoretic Problems for Benchmarking LLM Reasoning}
Graph-theoretic problems, such as shortest path, graph isomorphism test, and traveling salesman problem, are all classical problems in computer science, and can be used to benchmark LLM's reasoning capacity. Wang et al.~\cite{wang2023can} evaluate LLMs on graph-theoretic tasks and discover their weak reasoning abilities. Later, Tang et al.~\cite{tang2025grapharena} and Yuan et al.~\cite{yuan2025gracore} include more tasks, finding that even OpenAI o1-mini struggles with complex tasks. Wu et al.~\cite{wu2024grapheval2000} and Li et al.~\cite{li2024can} emphasize code-oriented problem. Xu et al.~\cite{xu2025graphomni} unifies varying graph types, encodings, and prompt styles for a comprehensive evaluation. G1~\cite{G1} proposes more tasks on larger graphs, and finetuning LLM with the dataset, leading to a small model with strong graph capacity.

\section{Graph for Enhancing LLM Capacity} 

The integration of LLMs and GNNs is a burgeoning field, with research exploring how each technology can enhance the other. While LLMs have been widely used to augment GNNs for traditional graph tasks like link prediction and node classification, the ability to process graphs and other structured data is also increasingly recognized as a core capacity of LLMs. This is particularly true for language tasks that have an implicit graph structure, such as navigating web pages, understanding code, and querying relational databases. In the following, we introduce how graphs, as a new modality, can be injected into LLMs and how such graph-structured data enhances LLMs' performance on certain tasks.

\subsection{Injecting Graphs into LLM Inputs}

A key challenge in combining LLMs with graphs is the fundamental mismatch between their data structures: LLMs process sequential text, while graphs are non-sequential and lack a natural order. To address this, researchers have explored several methods for injecting graph information into LLM inputs.

\paragraph{Graph as Text}
Initial approaches directly converted graph structures into text formats, like adjacency lists \cite{liu2023evaluating, guo2023gpt4graph, chen2024exploring}. While these methods have shown some promise, they often suffer from issues related to node and edge ordering and can struggle with large graphs due to the limited context windows of LLMs. Explorations with different encoding schemes \cite{fatemi2023talk} and linearization orders \cite{chu2024graphsos, das2024which} have generally resulted in only modest improvements. Another line of research proposes post-training LLMs using instruction tuning \cite{luo2024graphinstruct,ye2024language,G1} or preference tuning \cite{chen2024graphwiz,InstruGraph} on graph problems. These methods achieve good performance on problems related to basic graph structural properties.

\paragraph{Graph as Embedding}
A more effective approach involves using GNNs to generate rich node representation sequences that are then fed into LLMs. For example, Chai et al.~\cite{chai2023graphllm} used embeddings from Message Passing Neural Networks (MPNNs) to represent target nodes, enabling the LLM to answer basic structural questions. Similarly, Tang et al.~\cite{tang2024graphgpt} aligned both structural and textual inputs using GNNs and Transformers, achieving better accuracy by processing all node embeddings through a frozen pretrained LLM. G-Retriever~\cite{He2024GRetrieverRGK} and LlaGA~\cite{Wang2024LLaGALLI} extract textual information from graphs as context to boost the LLM's graph QA ability. Zhang et al.~\cite{Zhang2024GraphTranslatorAGG} use a frozen pretrained GNN to extract graph representations to feed to the LLM. Despite their success, these approaches can be limited by task-agnostic encoders, as seen in Qin et al.~\cite{qin2024disentangled}, which restrict their ability to transfer knowledge across different domains.

\paragraph{Merging GNN into LLM}
Recent efforts aim for a more profound integration by combining GNNs and Transformers. One example is GraphFormers~\cite{yang2023graphformers}, which proposed an architecture that combines GNNs and Transformers by iteratively encoding text and aggregating graph structures. While this approach improves text encoding for graph nodes, it overlooks the inclusion of task descriptions or language prompts. Building on this, Jin et al.~\cite{jin2023patton} adapted the GraphFormers architecture by refining its pretraining strategy to better capture text relationships between neighboring nodes. GOFA~\cite{Kong2024GOFAAGB} and CGM~\cite{CodeFuse} introduce graph information directly into the transformer layers, allowing their models to seamlessly process a mixture of graph and text.

{
\paragraph{Comparison between Different Injection methods}

We can compare these methods along three key dimensions: 
\begin{itemize}
    \item Changes to the base LLM model and training cost: In general, more substantial modifications to the base LLM tend to disrupt its pre-trained capabilities, necessitating additional training. TheGraph-as-Text approach makes no architectural changes and typically requires no training, allowing it to leverage closed-source, very large LLMs directly.Graph-as-Embedding usually involves training a graph encoder and sometimes fine-tuning the LLM, resulting in modest computational costs. In contrast,Merging GNN into LLM fundamentally alters the model architecture, requiring extensive pretraining; to our knowledge, this has so far only been feasible with relatively small LLMs (under 10 billion parameters). 
    \item Performance, which is highly task-dependent: for tasks involving simple graph structures but rich textual node features—such as question answering on small relational knowledge graphsGraph-as-Text often excels by fully exploiting the LLM’s strong language understanding. However, for tasks dominated by complex graph topology and weak or absent textual features—like molecular property predictionGraph-as-Embedding orMerged GNN-LLM architectures perform significantly better due to their explicit structural reasoning capabilities. 
    \item Inference cost and constraints on graph size, which are largely determined by how many tokens a graph consumes: inGraph-as-Text, each node or edge may require an entire sentence or paragraph, limiting practical use to graphs with only tens of tokens; inMerged GNN-LLM approaches, each node typically occupies one token, enabling handling of graphs with hundreds of nodes; meanwhile,Graph-as-Embedding methods can compress an entire graph—sometimes with thousands of nodes—into a single or fixed-size vector via the encoder, making them the most scalable for large-scale graph inputs.
\end{itemize}
}
\subsection{LLMs on Language Tasks with Graph Structure}

Graphs are increasingly used as a structural framework to help LLMs understand complex, non-linear information in various domains.

\paragraph{Tool Use}
In complex tool orchestration, frameworks like ToolNet~\cite{toolnet_2506} and ControlLLM~\cite{control_llm} use directed graphs to manage a vast collection of tools. These graphs explicitly model tool dependencies, allowing the LLM to efficiently navigate multi-step workflows by selecting the optimal next tool. This approach provides a crucial organizational structure that a simple list of tool descriptions lacks, enabling LLMs to scale to thousands of tools.

\paragraph{Code}
To comprehend the intricate structure of a codebase, LLMs can leverage graph representations. The Code Graph Model (CGM)~\cite{code_cgm} represents a code repository as a directed, heterogeneous graph, with nodes for code entities and edges for dependencies. This structural information is integrated into the LLM's attention mechanism, effectively allowing it to perform a form of graph-based message passing. Similarly, the LocAgent~\cite{loc_agent} framework uses a directed heterogeneous graph for code localization, enabling multi-hop reasoning by linking natural language queries to specific code elements.

\paragraph{Tabular Data}
Different graph structures are employed to preserve the relationships within and between tables. GraphOTTER~\cite{graph_otter} transforms a single table into a simple, undirected graph where each cell is a node and edges connect cells in the same row or column. This allows for a step-by-step reasoning process that filters out irrelevant information. For multi-table queries, the Schema Graph-Assisted Multi-table QA (SGAM)~\cite{sgam} framework uses a human-curated, directed schema graph where nodes are table columns and edges encode both intra-table and inter-table relationships. This explicitly provides schema links and join paths to the LLM, simplifying complex reasoning. Other approaches, such as the Table-Tree~\cite{table_tree} and Hybrid Graph~\cite{hybrid_graph}, decompose complex tables hierarchically or unify different data modalities into a single graph, respectively, to enhance reasoning and reduce token usage.

\section{Future Outlook}

\paragraph{Sparse LLM Architecture}


{
In the pursuit of larger models and improved computational efficiency, sparsity has been increasingly incorporated into architectural designs—and sparsity naturally induces a graph structure. For instance, the Mixture-of-Experts (MoE) architecture~\cite{moe_architecture} replaces the single large feed-forward network (FFN) in a standard Transformer layer with multiple smaller expert subnetworks. A lightweight, trainable component—called a router or gating network—dynamically selects which experts process each input token. This setup can be modeled as a bipartite graph: let layers and experts be nodes, and edges be connection between one layer to the selected expert. In early MoE models, this activation graph was simple, as only one expert was chosen per layer, giving each layer node degree one. Later work~\cite{meituan_longcat} extended this by allowing up to three experts per layer, yielding richer connectivity patterns and greater expressivity. Similarly, in the attention mechanism of LLMs, a score is computed between every pair of tokens; treating tokens as nodes and attention scores as weighted edges yields a dense graph in standard LLMs, where each token attends to all previous ones. To reduce complexity, sparse attention methods~\cite{NSA_deepseek} restrict attention to only a small subset of prior tokens. Notably, StreamingLLM~\cite{StreamingLLM} observed that in dense attention graphs, certain super nodes (tokens) exhibit very high degree, playing a critical role in maintaining context coherence. By preserving connections to these super nodes while pruning others, StreamingLLM achieves effective sparse attention with minimal performance loss. In summary, sparsity—and the graph structures it implies—is common in modern LLMs: we can either enrich the graph (e.g., by diversifying module interconnections) to enhance model expressivity, or prune it while preserving key structural properties (e.g., dominant attention pathways) to boost efficiency with acceptable performance degradation.
}

\paragraph{Brain-Inspired Knowledge Organization}


{
The human brain's memory is not a singular, monolithic storage unit but a complex, multi-stage process involving encoding, storage, and retrieval~\cite{human_memory}. It comprises transient short-term memory—supported by dynamic neuronal activity—and enduring long-term memory, with the hippocampus playing a crucial role in consolidating information from the former to the latter. This system is fundamentally sparse, localized, and abstract: rather than storing every sensory detail, humans typically recall high-level concepts and semantic associations—reminiscent of a knowledge graph~\cite{sparse_distributed_coding}. Recently, knowledge graphs have been integrated into AI agent systems to complement large language models (LLMs). For example, GraphRAG~\cite{GraphRAG} equips an LLM with a tool to query an external knowledge graph: the LLM generates a natural language query, and GraphRAG computes a combined score based on text similarity and graph structure to rank relevant entities or facts, returning the highest-scoring textual snippets from the graph. Beyond external retrieval, AriGraph~citep{AriGraph} goes further by organizing an agent’s past interactions and experiences into a dynamic graph. During operation, the agent can update this graph and retrieve information from it using natural language commands, effectively allowing the LLM to maintain short-term memory in its parameters while offloading long-term memory to the external knowledge graph. Some approaches even modify the LLM architecture itself to resemble a knowledge-graph-like “brain.” BriLLM~\cite{BriLLM}, for instance, uses a knowledge graph as its core architecture: each token type corresponds to a node, and semantic relationships are encoded as edges. During inference, input tokens activate their corresponding nodes, and signals propagate along edges to influence other nodes—mimicking associative recall. Although such novel architectures currently lag behind state-of-the-art LLMs in performance, they represent a promising direction toward more structured, interpretable, and memory-efficient models.

\paragraph{Comprehensive Evaluation of Graph Injection Methods}

A major gap in the current literature—-particularly in graph injection for large language models (LLMs)-—is the lack of standardized, large-scale empirical benchmarks. Most existing works are evaluated under heterogeneous conditions, which hinders direct comparison. These discrepancies span multiple dimensions: (1) base models, as different studies employ distinct LLMs; (2) pretraining data, which varies in scale, source, format, and whether it includes graph-structured prompts; (3) evaluation tasks, where although common benchmarks involve node classification, link prediction, and graph property prediction, most papers assess only a subset; (4) evaluation metrics, since many methods output text alone—excluding regression or classification metrics that require logits (e.g., AUROC)—while approaches without text generation capabilities cannot be evaluated on QA-style benchmarks; and (5) learning settings, which range from full supervision and fine-tuning to few-shot and zero-shot paradigms. Establishing unified evaluation protocols across these axes is crucial for systematic progress in the field.
}
\section{Acknowledgement}
This work is supported by National Natural Science Foundation of China (62550138).

\bibliographystyle{fcs}
\bibliography{ref}

\begin{biography}{FCS-251651-fig3} Xiyuan Wang is now a Ph.D. student at the Institute for Artificial Intelligence, Peking University, advised by Prof. Muhan Zhang.  His research focuses on vision generative model and graph neural network.
\end{biography}

\begin{biography}{FCS-251651-fig4} Yi Hu is now a Ph.D. student at the Institute for Artificial Intelligence, Peking University, advised by Prof. Muhan Zhang. She is dedicated to exploring the reasoning mechanisms of large language models (LLMs) and researching how to enhance the models’ reasoning capabilities to the level of human experts. 
\end{biography}

\begin{biography}{FCS-251651-fig5}
    Yanbo Wang is a Ph.D. student in Computer Science and Technology at Peking University, advised by Prof. Muhan Zhang. He is passionate about exploring a variety of topics, from using relational \& structured deep learning to understand relational data to leveraging language models or synthetic data for tabular data design and analysis. 
\end{biography}

\begin{biography}{FCS-251651-fig6}
    Chuan Shi is the Second-Class Professor of Beijing University of Posts and Telecommunications and the Changjiang Scholar Distinguished Professor of the Ministry of Education. His main research interests include Graph Machine Learning, Artificial Intelligence, and Scientific Intelligence.
\end{biography}

\begin{biography}{FCS-251651-fig7}
    Muhan Zhang is a tenure-track assistant professor and PhD advisor in the Institute for Artificial Intelligence of Peking University . His research focus on graph machine learning and large language models.
\end{biography}

\end{document}